\documentclass{article}

\usepackage[final]{acl}
\usepackage[utf8]{inputenc} 
\usepackage[T1]{fontenc}    
\usepackage{hyperref}       
\usepackage{url}            
\usepackage{booktabs}       
\usepackage{amsfonts}       
\usepackage{nicefrac}       
\usepackage{microtype}      
\usepackage{xcolor}         

\usepackage{amsmath}
\usepackage{amssymb}
\usepackage{amsthm}
\usepackage{mathtools}
\usepackage{bm}
\usepackage{xcolor}
\usepackage{accents}

\definecolor{blue}{RGB}{0,114,178}
\definecolor{red}{RGB}{213,80,20}
\definecolor{green}{RGB}{0,158,115}
\definecolor{purple}{RGB}{204,121,167}
\definecolor{orange}{RGB}{230,159, 0}
\definecolor{pink}{RGB}{204,121,167}

\newcommand{\D}{\mathcal{D}}
\newcommand{\Dtrain}{\D_{\mathrm{train}}}
\newcommand{\Dpool}{\D_{\mathrm{pool}}}

\newcommand{\x}{\mathbf{x}}

\newcommand{\y}{\mathrm{y}}

\usepackage{hyperref}
\usepackage{tabularx}
\usepackage{booktabs}
\usepackage{multirow}
\usepackage{adjustbox}
\usepackage{caption}
\usepackage{subcaption}
\usepackage{tipa} 

\usepackage{enumitem}
\setlist{nolistsep}
\setlist{nosep}
\usepackage{amssymb}
\usepackage{amsthm}
\usepackage{pifont} 
\usepackage{array}
\usepackage{comment}
\usepackage{amsmath}
\usepackage{mathtools}
\usepackage{algorithm}
\usepackage{algpseudocode}
\newcommand{\xmark}{\ding{55}}

\title{Advancing African-Accented English Speech Recognition: Epistemic Uncertainty-Driven Data Selection for Generalizable ASR Models}

\author{%
  Bonaventure F. P. Dossou\\
  McGill University, Mila Quebec AI Institute\\
  \texttt{bonaventure.dossou@mila.quebec}}

\begin{document}
\maketitle

\begin{abstract}
\label{abstract}
Accents play a pivotal role in shaping human communication, enhancing our ability to convey and comprehend messages with clarity and cultural nuance. While there has been significant progress in Automatic Speech Recognition (ASR), African-accented English ASR has been understudied due to a lack of training datasets, which are often expensive to create and demand colossal human labor. By combining several active learning paradigms and the core-set approach, we propose a new multi-round adaptation process that utilizes epistemic uncertainty to automate annotation, thereby significantly reducing associated costs and human labor. This novel method streamlines data annotation and strategically selects data samples that contribute most to model uncertainty, thereby enhancing training efficiency. We define a new U-WER metric to track model adaptation to hard accents. We evaluate our approach across several domains, datasets, and high-performing speech models. Our results show that our approach leads to a 27\% WER relative average improvement while requiring, on average, 45\% less data than established baselines. Our approach also improves out-of-distribution generalization for very low-resource accents, demonstrating its viability for building generalizable ASR models in the context of accented African ASR. We open-source the code 
\href{https://github.com/bonaventuredossou/active_learning_african_asr}{here}.
\end{abstract}

\section{Introduction}
\label{introduction}
Automatic Speech Recognition (ASR) is an active research area that powers voice assistant systems (VASs) like Siri and Cortana, enhancing daily communication \cite{kodish2018systematic,finley2018dictations,zapata2015assessing}. Despite this progress, no current VASs include African languages, which account for about 31\% of the world languages, and their unique accents (\cite{Ethnologue_Eberhard, tsvetkov2017opportunities}). This gap highlights the need for ASR systems that can effectively handle the linguistic diversity and complexity of African languages, particularly in critical applications such as healthcare. Due to the lack of representations of these languages and accents in training data, existing ASR systems often perform inadequately, even mispronouncing African names (\cite{olatunji23_interspeech}).

To address these challenges, our work focuses on adapting pre-trained speech models to transcribe African-accented English more accurately, characterized by unique intonations and pronunciations \cite{BENZEGHIBA2007763, hinsvark2021accented}. We use \textbf{epistemic uncertainty (EU)} \cite{kendall2017uncertainties} to guide the adaptation process by identifying gaps in model knowledge and prioritizing data for the model to learn from next. This is particularly beneficial in scenarios where data annotation is costly or time-consuming, as often seen in the African context \cite{badenhorst2019usefulness, badenhorst2017limitations, barnard2009asr, yemmene2019motivations, dichristofano2022performance, dossou2022afrolm, dossou2021okwugb}. EU also improves robustness and encourages exploration to mitigate inductive bias from underrepresented accents. Common approaches to compute EU include Monte Carlo Dropout (MC-Dropout) \cite{gal2016dropout} and Deep Ensembles \cite{deepensemble}, with the latter being more effective but computationally expensive. Due to resource constraints, we utilize MC-Dropout, which necessitates that models incorporate dropout components during pretraining.

We employ \textbf{Active Learning (AL)} techniques further to enhance the efficiency and effectiveness of model adaptation. AL leverages epistemic uncertainty to select the most informative data points from an unlabeled dataset for labeling, thereby improving model performance with fewer training instances. Common types of AL include Deep Bayesian Active Learning (DBAL) \cite{gal2017deep, houlsby2011bayesian} and Adversarial Active Learning (AAL) \cite{ducoffe2018adversarial}. AAL selects examples likely to be misclassified by the current model, refining it iteratively by challenging it with complex cases to enhance robustness. The core-set approach (CSA) \cite{sener2017active} is also related, as it selects a subset of the training data to ensure that a model trained on this subset performs comparably to one trained on the entire dataset, thereby addressing scalability and efficiency. A critical component of AL is the \textbf{acquisition function (AF)}, which determines the most informative samples from an unlabeled dataset for labeling. Key AFs include uncertainty sampling (US) \cite{liu2023understanding}, Bayesian Active Learning by Disagreement (BALD) \cite{gal2017deep}, and BatchBALD \cite{kirsch2019batchbald}. US targets data points with the highest model uncertainty. BALD maximizes the mutual information between model parameters and predictions. BatchBALD is an extension of BALD that selects multiple samples simultaneously but may choose redundant points. US is the least computationally expensive, making it ideal for efficient data labeling.

In this work, we leverage and combine DBAL, AAL, US, and CSA in the following way (in order): First, we integrate the CSA by leveraging smaller training subsets ($\sim 45\%$ smaller than the entire available training sets). Second, we utilize DBAL with MC-Dropout to apply dropout during both training and inference, thereby estimating the Bayesian posterior distribution. This allows us to practically and efficiently estimate EU in the models used \cite{gal2017deep} (see section \ref{epistemicu} for more details). Third, we use the estimated EU and integrate the idea of AAL using the US acquisition function.

We evaluate our approach across \textbf{several domains} (general, clinical, general+clinical aka \textit{both}), \textbf{several datasets} (AfriSpeech-200 \cite{Olatunji2023AfriSpeech200PA}), SautiDB \cite{afonja2021learning}, MedicalSpeech, CommonVoices English Accented Dataset \cite{commonvoice}, and \textbf{several high-performing speech models} (Wav2Vec2-XLSR-53 \cite{conneau2020unsupervised}, HuBERT-Large \cite{hsu2021hubert}, WavLM-Large \cite{chen2022wavlm}, and NVIDIA Conformer-CTC Large (en-US) \cite{gulati2020conformer}. \textbf{Our results show a 27\% Word Error Rate (WER) relative average improvement while requiring 45\% less data than established baselines.} We also adapt the standard WER to create an Uncertainty WER (U-WER) metric to track model adaptation to African accents.

The impact of our approach is substantial. It develops more robust, generalizable, and cost-efficient African-accented English ASR models, reducing dependency on large labeled datasets and enabling deployment in various real-world scenarios. Our results demonstrate improved generalization for out-of-distribution (OOD) cases, particularly for accents with limited resources, addressing specific challenges in African-accented automatic speech recognition (ASR). Additionally, by focusing on equitable representation in ASR training, our methodology promotes fairness in AI, ensuring technology serves users across diverse linguistic backgrounds without bias \cite{uncertainty_fairness_1, uncertainty_fairness_2, uncertainty_fairness_3}.
Our contributions are listed as follows:
\raggedbottom
\begin{itemize}
\item we combine DBAL, AAL, CSA, and EU to propose a novel way to adapt several high-performing pretrained speech models to build efficient African-accented English ASR models,
\item we evaluate our approach across several speech domains (clinical, general, \textit{both}), and African-accented speech datasets AfriSpeech-200 \cite{Olatunji2023AfriSpeech200PA}, SautiDB \cite{afonja2021learning}, MedicalSpeech, and CommonVoices English Accented Dataset \cite{commonvoice}, while providing domain and accent-specific analyses,
\item we define a new and simple metric called U-WER that allows us to measure and track how the variance of the model, across hard accents, changes over the adaptation process,
\item we show that our approach improves the relative average WER performance by 27\% while significantly reducing the required amount of labeled data (by $\sim$45\%),
\item we show, based on additional AL experiments, that our approach is also efficient in real-world settings where there are no gold transcriptions.
\end{itemize}

\section{Background and Related Works}
\subsection{Challenges for African-accented ASR}
State-of-the-art (SOTA) ASR technologies, powered by deep learning and neural network architectures like transformers, achieve high accuracy with Standard American English and major European languages. However, they often fail with African accents due to high variability in pronunciation and lack of quality speech data \cite{koenecke2020racial, das2021best}. This results in racial bias, poor performance, and potential social exclusion as speakers might alter their speech to be understood \cite{koenecke2020racial, koenecke21_spsc, chiu18_interspeech, mengesha2021don}. Enhancing Automatic Speech Recognition (ASR) for African languages is crucial for achieving equitable voice recognition, particularly in healthcare, education, and customer service. Solutions should focus on diversifying training datasets and developing robust modeling techniques tailored to the unique characteristics of these languages.
\subsection{Active Learning}
\label{active_learning}
AL aims to reduce the number of labeled training examples by automatically processing unlabeled examples and selecting the most informative ones, considering a given cost function, for a human to label. It is particularly effective when labeled data is scarce or expensive, optimizing the learning process by focusing on samples that most improve the model performance and generalization \cite{settles2009active, gal2017deep}. Several works have demonstrated its effectiveness and efficiency. An AL setup involves an unlabeled dataset $\Dpool = \{\x_i\}_{i = 1}^{n_{\mathrm{pool}}}$, a labeled training set $\Dtrain = \{\x_i, \y_i\}_{i = 1}^{n_{\mathrm{train}}}$, and a predictive model with likelihood $p_{w}(y|x)$ parameterized by $w \sim p(W|\Dtrain)$ ($W$ are the parameters of the model). The setup assumes the presence of an oracle to provide predictions $\y$ for all $x_{i} \in \Dpool$. After training, a batch of data $\{ \x^*_i \}_{i=1}^{b}$ is selected from $\Dpool$ based on its EU.

In \cite{al_speech_1}, AL was applied to a toy dataset of \textit{How May I Help You} recordings. Confidence scores were estimated for each word and used to compute the overall confidence score for the audio sample. This approach achieved competitive results using 27\% less data compared to the baseline. In \cite{al_speech_2}, the authors estimated confidence scores for each utterance using an online algorithm with the lattice output of a speech recognizer. The utterance scores were filtered through an informativeness function to select an optimal subset of training samples, reducing the labeled data needed for a given WER by over 60\%. \citet{accent_al} experimented with AL for accent adaptation in speech recognition. They adapted a source recognizer to the target accent by selecting a small, matched subset of utterances from a large, untranscribed, multi-accented corpus for human transcription. They employed a cross-entropy-based relevance measure in conjunction with uncertainty-based sampling. However, their experiments on Arabic and English accents showed worse performance compared to baselines while using more hours of recordings.
\begin{figure*}[tbh]
    \centering
    \includegraphics[width=\textwidth]{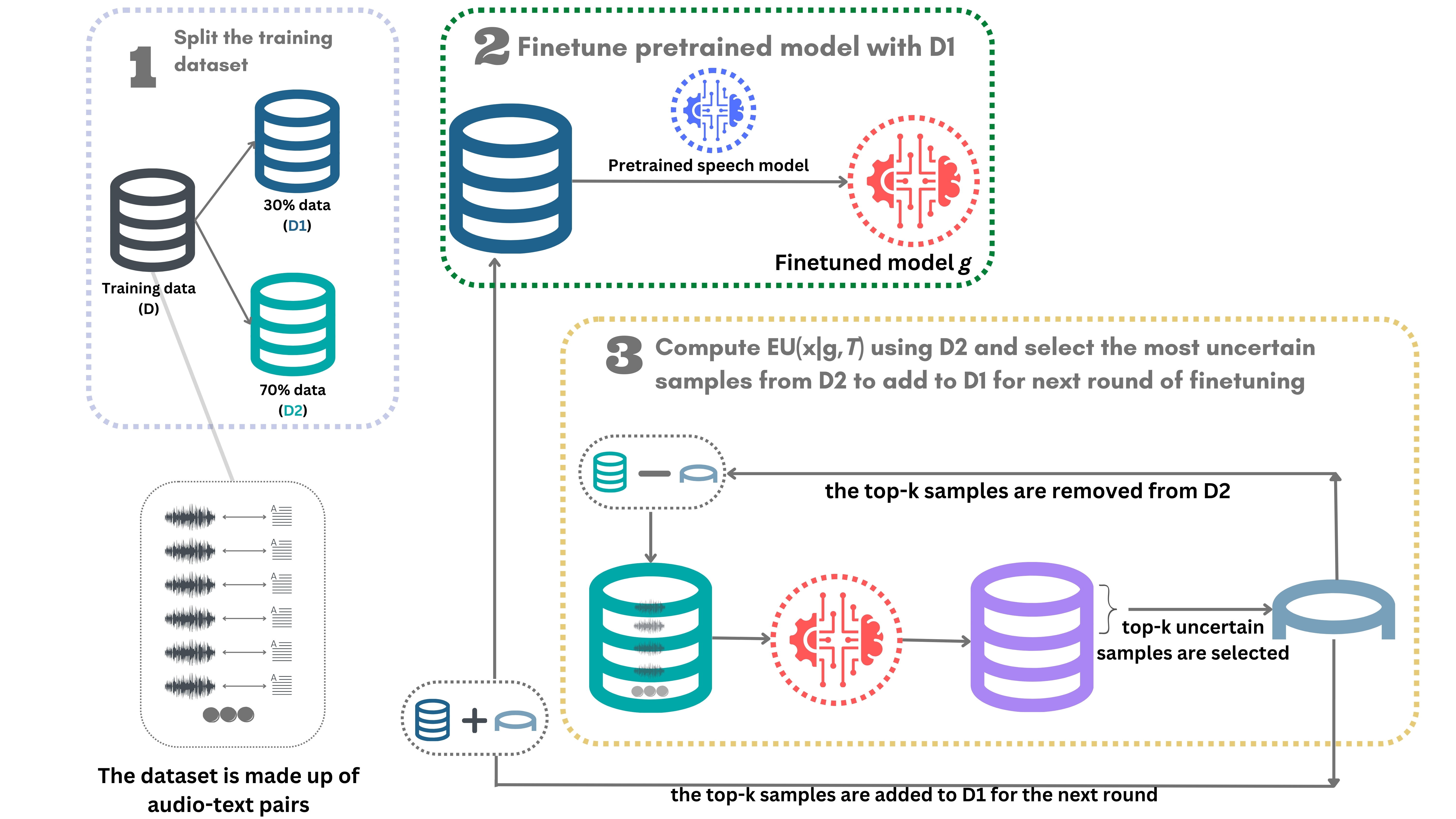}
\caption{Our adaptation pipeline involves several phases. Initially, the dataset is split into a training set ($D1 = \Dtrain^{*}$, 30\%) and a pool dataset ($D2 = \Dpool$, 70\%). In the iterative process between phases 2 and 3, $D1$ is used to finetune a pretrained model. The top-k samples are selected using defined strategies and added to $D1$ for the next round. For more details on the uncertainty selection strategy, see section \ref{epistemicu}.}
    \label{fig:my_label}
\end{figure*}

\section{Datasets and Methodology}
\subsection{Datasets}
\label{dataset}
We used the AfriSpeech-200 dataset \cite{Olatunji2023AfriSpeech200PA}, a 200-hour African-accented English speech corpus for clinical and general ASR. This dataset comprises over 120 African accents from five language families: Afro-Asiatic, Indo-European, Khoe-Kwadi (Hainum), Niger-Congo, and Nilo-Saharan, representing the diversity of African regional languages. It was crowd-sourced from over 2000 African speakers from 13 anglophone countries in sub-Saharan Africa and the US (see Table \ref{tab: dataset stats}).

To demonstrate the dataset-agnostic nature of our approach, we also explored three additional datasets: (1) \textbf{SautiDB} \cite{afonja2021sautidb}, Nigerian accent recordings with 919 audio samples at a 48kHz sampling rate, totaling 59 minutes; (2) \textbf{MedicalSpeech}\footnote{\url{https://www.kaggle.com/datasets/paultimothymooney/medical-speech-transcription-and-intent}}, containing 6,661 audio utterances of common medical symptoms, totaling 8 hours; and (3) \textbf{CommonVoices English Accented Dataset}, a subset of English Common Voice (version 10) \cite{commonvoice}, excluding western accents to focus on low-resource settings.

\begin{table}[h]
\centering
\caption{AfriSpeech-200 Dataset statistics}
\small
\begin{tabular}{l|l}
\hline 
\multicolumn{2}{c}{\textbf{AfriSpeech Dataset Statistics}}\\
\hline
Total duration & 200.91 hrs  \\
Total clips & 67,577  \\
Unique Speakers & 2,463  \\
Average Audio duration & 10.7 seconds  \\
\hline
\multicolumn{2}{c}{\textbf{Speaker Gender Ratios - \# Clip \%}}\\
\hline
Female & 57.11\%  \\
Male & 42.41\%  \\
Other/Unknown & 0.48\% \\
\hline
\multicolumn{2}{c}{\textbf{Speaker Age Groups - \# Clips}}\\
\hline
<18yrs & 1,264 (1.88\%) \\
19-25 & 36,728 (54.58\%)  \\
26-40 & 18,366 (27.29\%)  \\
41-55 & 10,374 (15.42\%) \\
>56yrs & 563 (0.84\%)  \\
\hline
\multicolumn{2}{c}{\textbf{Clip  Domain - \# Clips}}\\
\hline
Clinical & 41,765 (61.80\%)  \\
General & 25,812 (38.20\%) \\
\hline
\end{tabular}

\label{tab: dataset stats}
\end{table}

\begin{table}
\centering
\small
\caption{Dataset splits showing speakers, number of clips, and speech duration in Train/Dev/Test splits.}
\resizebox{\columnwidth}{!}{
\begin{tabular}{l|c|c|c|c}
\hline
\multicolumn{4}{c}{\textbf{AfriSpeech-200 Dataset Splits}}\\
\toprule
\textbf{Item} & \textbf{Train ($\Dtrain^{*}$)} & \textbf{Dev}  & \textbf{Test} & \textbf{AL Top-\textit{k}}\\
\hline
\# Speakers & 1466 & 247 & 750&\xmark \\
\#  Hours & 173.4 & 8.74 & 18.77&\xmark \\
\#  Accents & 71 & 45 & 108&\xmark \\
Avg secs/speaker & 425.81 & 127.32 & 90.08&\xmark\\
clips/speaker & 39.56 & 13.08 & 8.46&\xmark  \\
speakers/accent & 20.65 & 5.49 & 6.94&\xmark \\
secs/accent & 8791.96 & 698.82 & 625.55&\xmark \\
\# general domain & 21682 (*6504) & 1407 & 2723&2000 \\
\# clinical domain & 36318 (*10895) & 1824 & 3623&3500 \\
\# \textit{both} domain & 58000 (*17400) & 3221 & 6346&6500 \\
\hline
\end{tabular}
}
\label{tab:splits}
\end{table}

\subsection{Methodology}
\label{epistemicu}
\begin{algorithm}[!ht]
\caption{Selection of the best-generated transcript in Active Learning for an input Sample $x$}\label{alg:ul}
\begin{algorithmic}[1]
\State we generate the predictions $\hat{y}_{1},..,\hat{y}_{T}$ corresponding to each stochastic forward pass ($T$=10 in our experiments)
\State we define a list variable called wer\_list and a dictionary variable called wer\_target\_dict, respectively tracking all pairwise WERs and the average pairwise WER of each target prediction
\For{$\forall$ i,j $\in \{1,...,T\}$}
\State $\rightarrow$ $\hat{y}_{i}$ is set as target transcription
\State $\rightarrow$ target\_wer = list()
\For{for j $\neq$ i}
\State $w=WER(\hat{y}_{j}, \hat{y}_{i})$
\State wer\_list.append($w$)
\State target\_wer.append($w$)
\EndFor
\State $wer_{\hat{y}_{i}}$ = mean(target\_wer)
\State wer\_target\_dict[$\hat{y}_{i}$] $\leftarrow$ $wer_{\hat{y}_{i}}$
\EndFor
\State $\hat{y}_{best}$ = $\hat{y}_{i}$, such that wer\_target\_dict[$\hat{y}_{i}$] = min(wer\_target\_dict.values())
\State \textbf{return} ($p_{best}$, std(wer\_list))
\end{algorithmic}
\end{algorithm}
In our approach, to compute EU for a given input $x \in \Dpool$, we perform MC-Dropout to obtain multiple stochastic forward passes through a finetuned ASR model $g$ with likelihood $p_{w\sim p(\mathbf{W} \mid \Dtrain^{*})}(y|x)$ where $\mathbf{W}$ is the weights of $g$. Let $f$ be a function that computes the WER between the predicted and the target transcripts. Let $T$ be the number of stochastic forward passes. For each pass $t$, we apply dropout, obtain the output transcript, and compute the WER:
\[
f_{t} = f(y, \hat{y}_t); \hat{y}_t = g(\mathbf{W}, \tilde{x}_t); \tilde{x}_t = x \cdot \mathbf{M}_t
\]
where $\mathbf{M}_t$ is a binary mask matrix sampled independently for each pass. EU($x|g,T$) can then be estimated from the $T$ stochastic forward passes as follows:
\begin{equation}
\text{EU($x|g,T$)} = \sigma(f) = \sqrt{\frac{1}{T} \sum_{t=1}^{T} f_t^2 - \left( \frac{1}{T} \sum_{t=1}^{T} f_t \right)^2}
\end{equation}
The use of MC-Dropout requires models to have dropout components during training. This exclusion applies to some models, such as Whisper \cite{whisper}, which we still fine-tuned and evaluated as a baseline. We utilize four state-of-the-art pre-trained models: Wav2Vec2-XLSR-53, HuBERT-Large, WavLM-Large, and NVIDIA Conformer-CTC Large (en-US), referred to as Wav2Vec, HuBERT, WavLM, and Nemo, respectively.
\subsubsection{Uncertainty WER}
To handle diverse accents, we aim to reduce the EU of the models across hard accents after each adaptation round. We define a metric called \textit{U-WER} to track this. To compute U-WER(\textcolor{red}{$a$}) where \textcolor{red}{$a$} is a hard accent, we condition EU on \textcolor{red}{$a$}:
\begin{equation}
\text{EU($x|g,T, \textcolor{red}{a}$)} = \sigma(f_{\textcolor{red}{a}}) = \sqrt{\frac{1}{T} \sum_{t=1}^{T} f_{t,\textcolor{red}{a}}^2 - \left( \frac{1}{T} \sum_{t=1}^{T} f_{t,\textcolor{red}{a}} \right)^2}
\end{equation}
where $x_{\textcolor{red}{a}}$ is the audio sample with accent \textcolor{red}{a} and \[
f_{t,\textcolor{red}{a}} = f(y_{\textcolor{red}{a}}, \hat{y}_{t,\textcolor{red}{a}}); \hat{y}_{t,\textcolor{red}{a}} = g(\mathbf{W}, \tilde{x}_{t,\textcolor{red}{a}}); \tilde{x}_{t,\textcolor{red}{a}} = x_{\textcolor{red}{a}} \cdot \mathbf{M}_t
\]

Ideally, U-WER$\rightarrow$0. The rationale behind U-WER is that as beneficial data points are acquired, U-WER should decrease or remain constant, indicating increased robustness, knowledge, and performance, which is crucial for generalization. During AL, U-WER is computed using pairwise WER scores among predicted transcriptions, not gold transcriptions (see section \ref{experiments}). To select the best-generated transcript for unlabeled speech $x$, we follow Algorithm \ref{alg:ul}.
\begin{algorithm}[!ht]
\caption{Adaptation Round using Epistemic Uncertainty-based Selection}\label{alg:al}
\begin{algorithmic}[1]
\Require Pretrained Model $\mathcal{M}$, Training Dataset $\Dtrain^{*}$, Validation Dataset $\mathcal{D}_{Val}$, and Pool Dataset $\Dpool$
\State $\mathcal{N} \gets 3$\Comment{Number of Adaptation Rounds}
\State $T \gets 10$\Comment{Number of Stochastic Forward Passes}
\For{$k \gets 1$ to $\mathcal{N}$}
\State $g \gets$ Finetune $\mathcal{M}$ on $\Dtrain^{*}$ using $\mathcal{D}_{Val}$
\State $\mathcal{EUL} \gets \{\}$ \Comment{List of Uncertainty Scores}
\For{$x$ in $\Dpool$} \Comment{x is an audio sample}
\State EU$_{x}$ $\gets$ EU($x|g,T$) \Comment{Epistemic Uncertainty of $x$}
\State $\mathcal{EUL} \gets \mathcal{EUL} \cup \{(x, \text{EU}_{x})\}$
\EndFor
\State $topk \gets \{x_{1},...,x_{k}\}$ \Comment{Samples with highest $\mathcal{EU}$}
\State $\Dtrain^{*} \gets \Dtrain^{*} \cup topk$
\State $\Dpool \gets \Dpool \setminus topk$
\EndFor
\end{algorithmic}
\end{algorithm}
\subsection{Experimental Design}
\label{experiments}

To work within our framework, we define the following selection strategies:
\begin{itemize}
\item \textbf{random}: Randomly selects audio samples from $\Dpool$.
\item \textbf{EU-Most}: Selects the most uncertain audio samples from $\Dpool$ to add to $\Dtrain$.
\item \textbf{AL-EU-Most}: Combines AL with the \textbf{EU-Most} strategy to finetune the pretrained model.
\end{itemize}
We also define \textbf{standard fine-tuning (SFT)} as baseline using all available data for finetuning. In SFT, $\Dpool$ is empty. While running the defined strategies in our framework, we \textbf{impose data constraints, not exceeding 60-65\% of the initial dataset after all adaptation rounds.} $\Dtrain^{*}$ is 30\% of $\Dtrain$, and $\Dpool$ is 70\% of $\Dtrain$. This simulates realistic scenarios where not all data might be available, testing the approach's robustness and efficiency under constraints. The number of samples in $\Dtrain^{}$ and $\Dpool$ is based on available training examples for each domain (see Tables \ref{tab:splits}, \ref{table:otherdatasets}, and Appendix \ref{sec:para}).

Our EU-based pipeline is shown in Figure \ref{fig:my_label} and Algorithm \ref{alg:al}. In each adaptation round, we use a finetuned model and a selection strategy to choose samples from $\Dpool$ to add to $\Dtrain^{*}$. During AL experiments, we consider samples from $\Dpool$ as unlabeled: (1) using MC-Dropout, we obtain $n=10$ different input representations per audio sample to get $n$ different transcripts; (2) we then learn to select the best-generated transcription as the target transcription according to Algorithm \ref{alg:ul}.

Our experiments aim to answer the following research questions:
\begin{enumerate}
\item how does the pretrained ASR model adapt to a set of African accents across adaptation rounds and domains?
\item which selection strategy (\textbf{EU-most} or \textbf{random}) works better, and for which domain(s)?
\item which domain(s) help the model perform better, and how does the model perform (in terms of uncertainty) across the domain(s)?
\item what is the impact of EU-based selection on the model's efficiency in low-resource data scenarios?
\item is uncertainty-based selection, model, and dataset agnostic?
\end{enumerate}

U-WER will answer question 4. To answer question 5, we evaluated our approach with three additional pretrained models (Nemo, WavLM, and Hubert) and across three external datasets (SautiDB, CommonVoices English Accented Dataset, and MedicalSpeech). For consistency and better visualization, we considered the top-10 (in terms of frequency) accents across three adaptation rounds and both selection strategies to answer questions 1-4. For very low-resource settings, we considered the five accents with the least recording hours.

For our experiments, we utilized six RTX 8000 GPUs and four A100 GPUs. Training and evaluation were conducted over a period of one month. Our models have approximately 311 million trainable parameters. Each audio sample was normalized and processed at a 16 kHz sample rate. We used default parameters from the HuggingFace library for each pretrained model.
\begin{table*}
\caption{\label{table:results}
We utilized Wav2Vec to conduct initial experiments across various domains and strategies, aiming to identify the optimal selection strategy. Models marked with ** are used to demonstrate that our algorithm is model agnostic, utilizing the \textbf{EU-Most} selection strategy, which has been proven to be the most effective. Our AL experiments also use this strategy. Wav2Vec, using the \textbf{random} strategy, scored 0.1111, 0.3571, and 0.1666 for the general, clinical, and \textit{both} domains, respectively. We omit \textbf{random} results to enhance readability.
}
\small
\centering
\resizebox{\textwidth}{!}{
\renewcommand{\arraystretch}{1.9}
\begin{tabular}{l||ccc||ccc||ccc}
\hline
\multirow{2}{*}{\textbf{Model}} & \multicolumn{3}{c}{\textbf{General}} & \multicolumn{3}{c}{\textbf{Clinical}}& \multicolumn{3}{c}{\textbf{\textit{Both}}}\\
 & \textbf{Baseline} & \textbf{EU-Most} & \textbf{AL-EU-Most} & \textbf{Baseline} & \textbf{EU-Most} & \textbf{AL-EU-Most} & \textbf{Baseline} & \textbf{EU-Most} & \textbf{AL-EU-Most}  \\
\hline
Wav2vec & 0.2360 \cite{Olatunji2023AfriSpeech200PA} & \textbf{0.1011} & 0.1059 & 0.3080 \cite{Olatunji2023AfriSpeech200PA}& \textbf{0.2457}& 0.2545 & 0.2950 \cite{Olatunji2023AfriSpeech200PA} & \textbf{0.1266} & \textbf{0.1309} \\

**Hubert & \textbf{0.1743} & 0.1901 & 0.1887 & 0.2907& \textbf{0.2594} & 0.2709 & \textbf{0.2365} & 0.2453 & 0.2586 \\

**WavLM & 0.1635 & \textbf{0.1576} & 0.1764 & 0.3076 & \textbf{0.2313} & 0.2537 & 0.2047 & \textbf{0.1897} & 0.1976 \\
**Nemo & 0.2824& \textbf{0.1765} &0.1815 & 0.2600 & \textbf{0.2492} & 0.2526 & 0.3765 & \textbf{0.2576} & 0.2610\\
\hline
Average Performance & 0.2141&\textbf{0.1563} & 0.1631 & 0.2916 & \textbf{0.2464} & 0.2579 & 0.2782 & \textbf{0.2043} &  0.2120\\
Whisper-Medium & 0.2806& - & - & 0.3443 & - & - & 0.3116 & - & - \\
\bottomrule
\end{tabular}
}
\end{table*}

\begin{figure*}[!ht]
  \subfloat[]{
	\begin{minipage}[c][1\width]{
	   0.5\textwidth}
	   \centering
	   \includegraphics[width=1.2\textwidth]{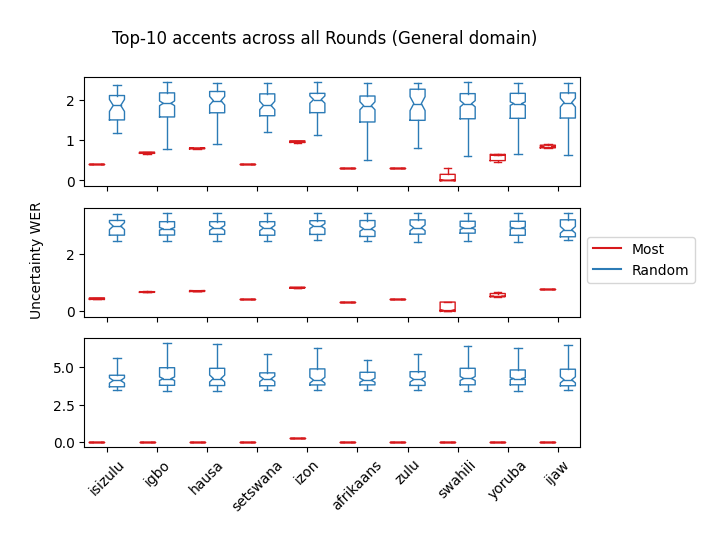}
	\end{minipage}}
 \hfill	
  \subfloat[]{
	\begin{minipage}[c][1\width]{
	   0.5\textwidth}
	   \centering   \includegraphics[width=1.2\textwidth]{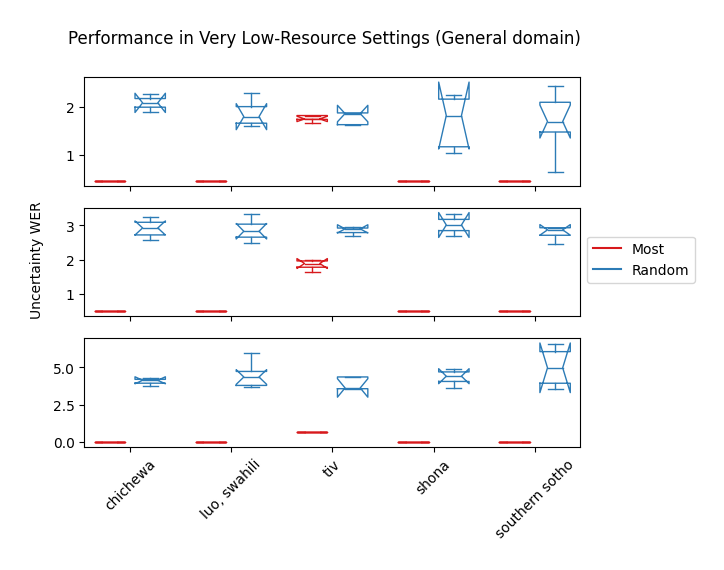}
	\end{minipage}}
\caption{WER Performance on Accents from General Domain}
\label{wer_general}
\end{figure*}

\begin{figure*}[!ht]
  \subfloat[]{
	\begin{minipage}[c][1\width]{
	   0.5\textwidth}
	   \centering
	   \includegraphics[width=1.2\textwidth]{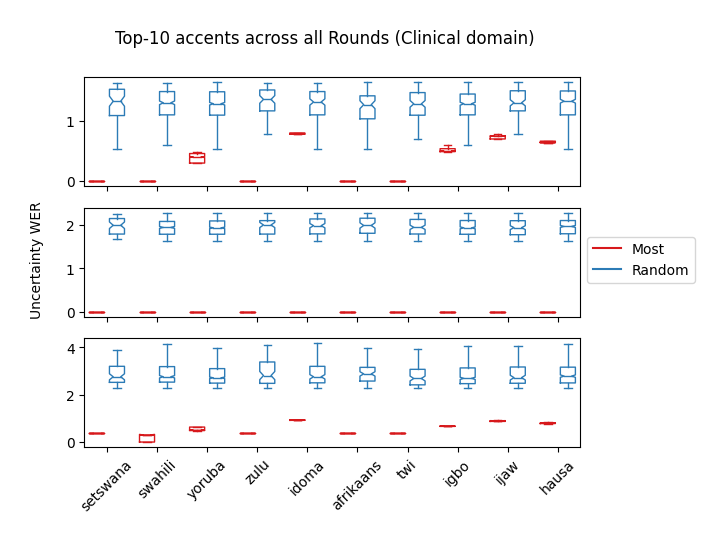}
	\end{minipage}}
 \hfill	
  \subfloat[]{
	\begin{minipage}[c][1\width]{
	   0.5\textwidth}
	   \centering   \includegraphics[width=1.2\textwidth]{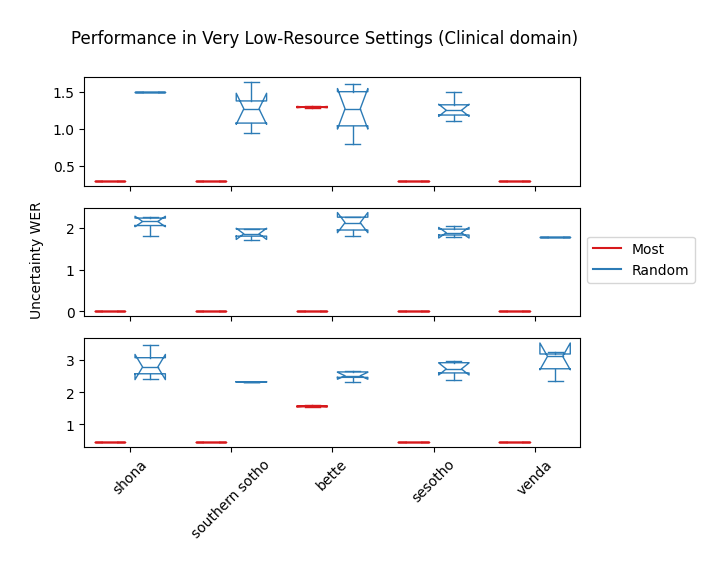}
	\end{minipage}}
\caption{WER Performance on Accents from Clinical Domain}
\label{wer_clinical}
\end{figure*}

\begin{figure*}[!ht]
  \subfloat[]{
	\begin{minipage}[c][1\width]{
	   0.5\textwidth}
	   \centering
	   \includegraphics[width=1.2\textwidth]{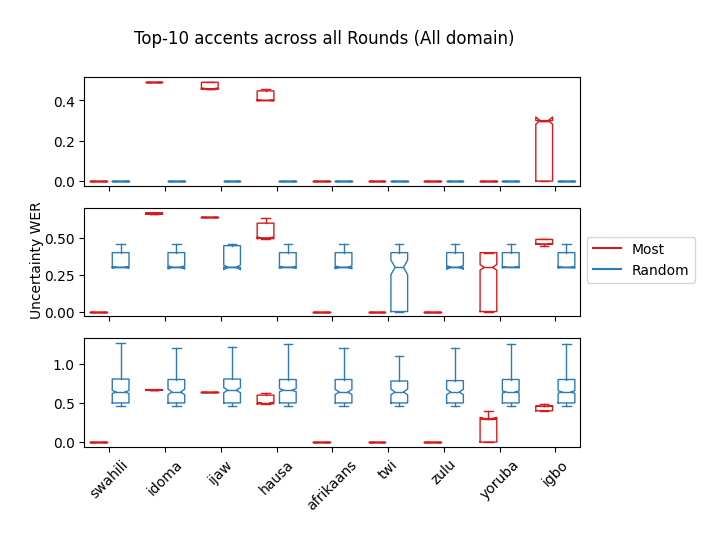}
	\end{minipage}}
 \hfill	
  \subfloat[]{
	\begin{minipage}[c][1\width]{
	   0.5\textwidth}
	   \centering   \includegraphics[width=1.2\textwidth]{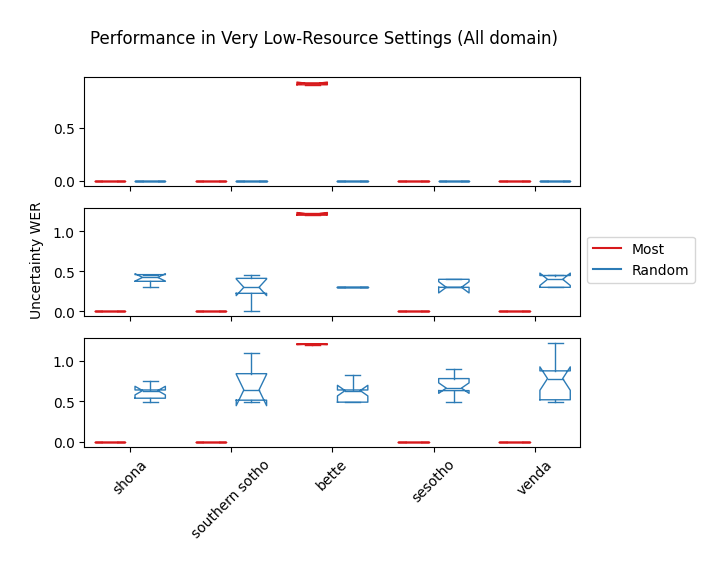}
	\end{minipage}}
\caption{WER Performance on Accents from Clinical+General (\textit{Both}) Domain}
\label{wer_all}
\end{figure*}

\begin{table*}[!ht]
\small
\caption{\label{table:otherdatasets} WER Evaluation Results on External Datasets, with $\alpha \in [0.60, 0.65]$ as described in Section \ref{dataset} and on Figure \ref{fig:my_label}. We observe an improvement in WER using our approach across all datasets, indicating that our algorithm is dataset-agnostic.
}
\begin{center}
\resizebox{\textwidth}{!}{
\renewcommand{\arraystretch}{1.9}
\begin{tabular}{ccccccccc}
\toprule
\multirow{2}{*}{\textbf{Dataset}} & \multicolumn{4}{c}{\textbf{Split and Size for our approach}} & \multirow{2}{*}{\textbf{Finetuning Epochs}}& \textbf{Baseline} & \textbf{EU-Most}\\
 & \textbf{$\Dtrain^{*}$} & \textbf{$\Dpool$} & \textbf{Top-\textit{k}}& \textbf{Test} & &\textbf{($\Dtrain$)}&\textbf{($\Dtrain^{*}$ + $\alpha\Dpool$)}\\
\midrule
SautiDB \cite{afonja2021sautidb}&234&547& 92& 138&50&0.50&\textbf{0.12}\\
MedicalSpeech&1598&3730&1333& 622&5&0.30&\textbf{0.28}\\
CommonVoices English Accented Dataset (v10.0) \cite{commonvoice} &26614&62100& 10350&232&5&0.50& \textbf{0.22}\\
\hline
Average& \xmark & \xmark & \xmark & \xmark& \xmark&0.43&\textbf{0.20}\\
\bottomrule
\end{tabular}
}
\end{center}
\end{table*}

\section{Results and Discussion}
\label{res_discussion}

To assess the performance improvement for each domain, we compute the relative average improvement \[\text{RIA}_{wer, d} = \left( \frac{b_{wer}^{d} - s_{wer}^{d}}{b_{wer}^{d}} \right) \times 100\%\] where $b_{wer}^{d}$ and $s_{wer}^{d}$ are the average WER respectively of the baseline, and the best selection strategy, in a domain $d \in \{general,clinical,both\}$. A higher percentage reflects a higher improvement in our approach.

Table \ref{table:results} shows the results of our experiments, indicating that our uncertainty-based selection approach significantly outperforms the baselines across \textbf{all models, domains, and datasets: general (27.00\%), clinical (15.51\%), and \textit{both} (26.56\%)}. Our approach also surpasses Whisper-Medium (\cite{Olatunji2023AfriSpeech200PA, radford2023robust}), demonstrating the importance of epistemic uncertainty in ASR for low-resource languages. The \textbf{EU-Most} selection strategy proves to be the most effective across all domains due to the model's exposure to highly uncertain samples, enhancing robustness and performance. However, performance disparities between the general and clinical domains are noted, likely due to the complexity of the clinical sample. These findings confirm \textbf{EU-Most} as the superior selection strategy, as detailed in the results and illustrated in Figures \ref{wer_general}, \ref{wer_clinical}, and \ref{wer_all}. This answers question 2.

To identify the best learning signals within a diverse dataset characterized by various accents, speaker traits, genders, and ages, we analyzed the top-k uncertain accents using the \textbf{EU-Most} selection strategy. Our findings, illustrated in Figures \ref{wer_general}, \ref{wer_clinical}, and \ref{wer_all}, show that the top-10 accents (most represented in recording hours) remained consistently challenging across all rounds of analysis (refer to Figures \ref{wer_general}, \ref{wer_clinical}, \ref{wer_all} and Tables \ref{tab:Country Stats}, \ref{tab:dataset_accent_stats1}, and \ref{tab:dataset_accent_stats2}). These accents, characterized by high linguistic richness and variability, facilitate model learning and improve performance over time. We positively answer questions 1 and 3, confirming that the model adapts effectively to the beneficial accents from all domains. This demonstrates that the model adapts qualitatively and quantitatively well to the beneficial accents and benefits from all domains. Figures \ref{wer_general} (b), \ref{wer_clinical} (b), and \ref{wer_all} (b) also affirm positive outcomes for question 4, showing consistent improvement or stable performance on low-resource accents. This highlights the relevance of our approach in addressing the challenges associated with the limited resource availability typical of many African languages and dialects.

To demonstrate the agnostic aspect of our approach, we evaluated it using three additional pre-trained models (Hubert, WavLM, and Nemo) and three datasets containing accented speech in general and clinical domains, employing only the \textbf{EU-Most} selection strategy. The results, shown in Tables \ref{table:results} and \ref{table:otherdatasets}, indicate that our uncertainty-based adaptation approach consistently outperforms baselines. This confirms that our approach applies to any model architecture and dataset, allowing us to answer question 5 positively.
\raggedbottom
\section{Conclusion}\label{sec:conc}
We combined several AL paradigms, the CSA, and the EU to create a novel multi-round adaptation process for high-performing pretrained speech models, aiming to build efficient African-accented English ASR models. We introduced the U-WER metric to track model adaptation to intricate accents. Our experiments demonstrated a remarkable 27\% WER ratio improvement while reducing the data required for effective training by approximately 45\% compared to existing baselines. This reflects the efficiency and potential of our approach to lower the barriers to ASR technologies in underserved regions significantly. Our method enhances model robustness and generalization across various domains, datasets, and accents, which are crucial for scalable ASR systems. This also helps mitigate bias in ASR technologies, promoting more inclusive and fair AI applications.
\section{Limitations}
\label{sec:limit}
In discussing trade-offs (Section \ref{res_discussion}), we noted that while our approach enhances performance, particularly with linguistically rich accents, a stopping criterion is essential for complex domains like the \textbf{clinical} one to balance adaptation rounds with the pool size. With better resources, we would consider implementing Deep Ensembles (\cite{deepensemble}) as an alternative to our current MC-Dropout method for estimating epistemic uncertainty and leveraging other acquisition functions (such as BALD, BatchBALD) highlighted in this work.
\newpage
\section{Acknowledgments}
The authors are very grateful to Prof. Ines Arous for her help in reviewing the manuscript and suggesting \textbf{very important} changes to improve the quality of the paper. The author acknowledges the support of the Mila Quebec AI Institute for computing resources.

The authors acknowledge Intron Health for providing the Afrispeech-200 dataset. The authors are grateful to Atnafu Lambebo Tonja, Chris Chinenye Emezue, Tobi Olatunji, Naome A Etori, Salomey Osei, Tosin Adewumi, and Sahib Singh for their help in the early stage of this project.

\bibliography{iclr2025_conference}

\begin{thebibliography}{46}
\providecommand{\natexlab}[1]{#1}
\providecommand{\url}[1]{\texttt{#1}}
\expandafter\ifx\csname urlstyle\endcsname\relax
  \providecommand{\doi}[1]{doi: #1}\else
  \providecommand{\doi}{doi: \begingroup \urlstyle{rm}\Url}\fi

\bibitem[Afonja et~al.(2021{\natexlab{a}})Afonja, Mbataku, Malomo, Okubadejo, Francis, Nwadike, and Orife]{afonja2021sautidb}
T.~Afonja, C.~Mbataku, A.~Malomo, O.~Okubadejo, L.~Francis, M.~Nwadike, and I.~Orife.
\newblock Sautidb: Nigerian accent dataset collection, 2021{\natexlab{a}}.

\bibitem[Afonja et~al.(2021{\natexlab{b}})Afonja, Mudele, Orife, Dukor, Francis, Goodness, Azeez, Malomo, and Mbataku]{afonja2021learning}
T.~Afonja, O.~Mudele, I.~Orife, K.~Dukor, L.~Francis, D.~Goodness, O.~Azeez, A.~Malomo, and C.~Mbataku.
\newblock Learning nigerian accent embeddings from speech: preliminary results based on sautidb-naija corpus.
\newblock \emph{arXiv preprint arXiv:2112.06199}, 2021{\natexlab{b}}.

\bibitem[Ardila et~al.(2019)Ardila, Branson, Davis, Henretty, Kohler, Meyer, Morais, Saunders, Tyers, and Weber]{commonvoice}
R.~Ardila, M.~Branson, K.~Davis, M.~Henretty, M.~Kohler, J.~Meyer, R.~Morais, L.~Saunders, F.~M. Tyers, and G.~Weber.
\newblock Common voice: A massively-multilingual speech corpus.
\newblock \emph{arXiv preprint arXiv:1912.06670}, 2019.

\bibitem[Badenhorst and De~Wet(2017)]{badenhorst2017limitations}
J.~Badenhorst and F.~De~Wet.
\newblock The limitations of data perturbation for asr of learner data in under-resourced languages.
\newblock In \emph{2017 Pattern Recognition Association of South Africa and Robotics and Mechatronics (PRASA-RobMech)}, pages 44--49. IEEE, 2017.

\bibitem[Badenhorst and De~Wet(2019)]{badenhorst2019usefulness}
J.~Badenhorst and F.~De~Wet.
\newblock The usefulness of imperfect speech data for asr development in low-resource languages.
\newblock \emph{Information}, 10\penalty0 (9):\penalty0 268, 2019.

\bibitem[Barnard et~al.(2009)Barnard, Davel, and Van~Heerden]{barnard2009asr}
E.~Barnard, M.~Davel, and C.~Van~Heerden.
\newblock Asr corpus design for resource-scarce languages.
\newblock ISCA, 2009.

\bibitem[Benzeghiba et~al.(2007)Benzeghiba, {De Mori}, Deroo, Dupont, Erbes, Jouvet, Fissore, Laface, Mertins, Ris, Rose, Tyagi, and Wellekens]{BENZEGHIBA2007763}
M.~Benzeghiba, R.~{De Mori}, O.~Deroo, S.~Dupont, T.~Erbes, D.~Jouvet, L.~Fissore, P.~Laface, A.~Mertins, C.~Ris, R.~Rose, V.~Tyagi, and C.~Wellekens.
\newblock Automatic speech recognition and speech variability: A review.
\newblock \emph{Speech Communication}, 49\penalty0 (10):\penalty0 763--786, 2007.
\newblock ISSN 0167-6393.
\newblock \doi{https://doi.org/10.1016/j.specom.2007.02.006}.
\newblock URL \url{https://www.sciencedirect.com/science/article/pii/S0167639307000404}.
\newblock Intrinsic Speech Variations.

\bibitem[Chen et~al.(2022)Chen, Wang, Chen, Wu, Liu, Chen, Li, Kanda, Yoshioka, Xiao, et~al.]{chen2022wavlm}
S.~Chen, C.~Wang, Z.~Chen, Y.~Wu, S.~Liu, Z.~Chen, J.~Li, N.~Kanda, T.~Yoshioka, X.~Xiao, et~al.
\newblock Wavlm: Large-scale self-supervised pre-training for full stack speech processing.
\newblock \emph{IEEE Journal of Selected Topics in Signal Processing}, 16\penalty0 (6):\penalty0 1505--1518, 2022.

\bibitem[Chiu et~al.(2018)Chiu, Tripathi, Chou, Co, Jaitly, Jaunzeikare, Kannan, Nguyen, Sak, Sankar, Tansuwan, Wan, Wu, and Zhang]{chiu18_interspeech}
C.-C. Chiu, A.~Tripathi, K.~Chou, C.~Co, N.~Jaitly, D.~Jaunzeikare, A.~Kannan, P.~Nguyen, H.~Sak, A.~Sankar, J.~Tansuwan, N.~Wan, Y.~Wu, and X.~Zhang.
\newblock {Speech Recognition for Medical Conversations}.
\newblock In \emph{Proc. Interspeech 2018}, pages 2972--2976, 2018.
\newblock \doi{10.21437/Interspeech.2018-40}.

\bibitem[Conneau et~al.(2020)Conneau, Baevski, Collobert, rahman Mohamed, and Auli]{conneau2020unsupervised}
A.~Conneau, A.~Baevski, R.~Collobert, A.~rahman Mohamed, and M.~Auli.
\newblock Unsupervised cross-lingual representation learning for speech recognition.
\newblock \emph{INTERSPEECH}, 2020.
\newblock \doi{10.21437/interspeech.2021-329}.

\bibitem[Das et~al.(2021)Das, Bodapati, Sunkara, Srinivasan, and Chau]{das2021best}
N.~Das, S.~Bodapati, M.~Sunkara, S.~Srinivasan, and D.~H. Chau.
\newblock Best of both worlds: Robust accented speech recognition with adversarial transfer learning.
\newblock \emph{arXiv preprint arXiv:2103.05834}, 2021.

\bibitem[DiChristofano et~al.(2022)DiChristofano, Shuster, Chandra, and Patwari]{dichristofano2022performance}
A.~DiChristofano, H.~Shuster, S.~Chandra, and N.~Patwari.
\newblock Performance disparities between accents in automatic speech recognition.
\newblock \emph{arXiv preprint arXiv:2208.01157}, 2022.

\bibitem[Dossou and Emezue(2021)]{dossou2021okwugb}
B.~F. Dossou and C.~C. Emezue.
\newblock Okwugb$\backslash$'e: End-to-end speech recognition for fon and igbo.
\newblock \emph{arXiv preprint arXiv:2103.07762}, 2021.

\bibitem[Dossou et~al.(2022)Dossou, Tonja, Yousuf, Osei, Oppong, Shode, Awoyomi, and Emezue]{dossou2022afrolm}
B.~F. Dossou, A.~L. Tonja, O.~Yousuf, S.~Osei, A.~Oppong, I.~Shode, O.~O. Awoyomi, and C.~C. Emezue.
\newblock Afrolm: A self-active learning-based multilingual pretrained language model for 23 african languages.
\newblock \emph{arXiv preprint arXiv:2211.03263}, 2022.

\bibitem[Ducoffe and Precioso(2018)]{ducoffe2018adversarial}
M.~Ducoffe and F.~Precioso.
\newblock Adversarial active learning for deep networks: a margin based approach.
\newblock \emph{arXiv preprint arXiv:1802.09841}, 2018.

\bibitem[Eberhard et~al.(2019)Eberhard, Simons, and Fennig]{Ethnologue_Eberhard}
D.~Eberhard, G.~Simons, and C.~Fennig.
\newblock \emph{Ethnologue: Languages of the World, 22nd Edition}.
\newblock 02 2019.

\bibitem[Finley et~al.(2018)Finley, Salloum, Sadoughi, Edwards, Robinson, Axtmann, Brenndoerfer, Miller, and Suendermann-Oeft]{finley2018dictations}
G.~Finley, W.~Salloum, N.~Sadoughi, E.~Edwards, A.~Robinson, N.~Axtmann, M.~Brenndoerfer, M.~Miller, and D.~Suendermann-Oeft.
\newblock From dictations to clinical reports using machine translation.
\newblock In \emph{Proceedings of the 2018 Conference of the North American Chapter of the Association for Computational Linguistics: Human Language Technologies, Volume 3 (Industry Papers)}, pages 121--128, 2018.

\bibitem[Gal and Ghahramani(2016)]{gal2016dropout}
Y.~Gal and Z.~Ghahramani.
\newblock Dropout as a bayesian approximation: Representing model uncertainty in deep learning.
\newblock In \emph{international conference on machine learning}, pages 1050--1059. PMLR, 2016.

\bibitem[Gal et~al.(2017)Gal, Islam, and Ghahramani]{gal2017deep}
Y.~Gal, R.~Islam, and Z.~Ghahramani.
\newblock Deep {B}ayesian active learning with image data.
\newblock In D.~Precup and Y.~W. Teh, editors, \emph{Proceedings of the 34th International Conference on Machine Learning}, volume~70 of \emph{Proceedings of Machine Learning Research}, pages 1183--1192. PMLR, 06--11 Aug 2017.
\newblock URL \url{https://proceedings.mlr.press/v70/gal17a.html}.

\bibitem[Gulati et~al.(2020)Gulati, Qin, Chiu, Parmar, Zhang, Yu, Han, Wang, Zhang, Wu, et~al.]{gulati2020conformer}
A.~Gulati, J.~Qin, C.-C. Chiu, N.~Parmar, Y.~Zhang, J.~Yu, W.~Han, S.~Wang, Z.~Zhang, Y.~Wu, et~al.
\newblock Conformer: Convolution-augmented transformer for speech recognition.
\newblock \emph{arXiv preprint arXiv:2005.08100}, 2020.

\bibitem[Hakkani-Tür et~al.(2002)Hakkani-Tür, Riccardi, and Gorin]{al_speech_1}
D.~Hakkani-Tür, G.~Riccardi, and A.~Gorin.
\newblock Active learning for automatic speech recognition.
\newblock In \emph{2002 IEEE International Conference on Acoustics, Speech, and Signal Processing}, volume~4, pages IV--3904--IV--3907, 2002.
\newblock \doi{10.1109/ICASSP.2002.5745510}.

\bibitem[Hinsvark et~al.(2021)Hinsvark, Delworth, Rio, McNamara, Dong, Westerman, Huang, Palakapilly, Drexler, Pirkin, Bhandari, and Jette]{hinsvark2021accented}
A.~Hinsvark, N.~Delworth, M.~D. Rio, Q.~McNamara, J.~Dong, R.~Westerman, M.~Huang, J.~Palakapilly, J.~Drexler, I.~Pirkin, N.~Bhandari, and M.~Jette.
\newblock Accented speech recognition: A survey, 2021.

\bibitem[Houlsby et~al.(2011)Houlsby, Husz{\'a}r, Ghahramani, and Lengyel]{houlsby2011bayesian}
N.~Houlsby, F.~Husz{\'a}r, Z.~Ghahramani, and M.~Lengyel.
\newblock Bayesian active learning for classification and preference learning.
\newblock \emph{arXiv preprint arXiv:1112.5745}, 2011.

\bibitem[Hsu et~al.(2021)Hsu, Bolte, Tsai, Lakhotia, Salakhutdinov, and Mohamed]{hsu2021hubert}
W.-N. Hsu, B.~Bolte, Y.-H.~H. Tsai, K.~Lakhotia, R.~Salakhutdinov, and A.~Mohamed.
\newblock Hubert: Self-supervised speech representation learning by masked prediction of hidden units.
\newblock \emph{IEEE/ACM Transactions on Audio, Speech, and Language Processing}, 29:\penalty0 3451--3460, 2021.

\bibitem[Kendall and Gal(2017)]{kendall2017uncertainties}
A.~Kendall and Y.~Gal.
\newblock What uncertainties do we need in bayesian deep learning for computer vision?
\newblock \emph{Advances in Neural Information Processing Systems}, 30:\penalty0 5574--5584, 2017.

\bibitem[Kirsch et~al.(2019)Kirsch, Amersfoort, and Gal]{kirsch2019batchbald}
A.~Kirsch, J.~v. Amersfoort, and Y.~Gal.
\newblock \emph{BatchBALD: efficient and diverse batch acquisition for deep Bayesian active learning}.
\newblock Curran Associates Inc., Red Hook, NY, USA, 2019.

\bibitem[Kodish-Wachs et~al.(2018)Kodish-Wachs, Agassi, Kenny~III, and Overhage]{kodish2018systematic}
J.~Kodish-Wachs, E.~Agassi, P.~Kenny~III, and J.~M. Overhage.
\newblock A systematic comparison of contemporary automatic speech recognition engines for conversational clinical speech.
\newblock In \emph{AMIA Annual Symposium Proceedings}, volume 2018, page 683. American Medical Informatics Association, 2018.

\bibitem[Koenecke(2021)]{koenecke21_spsc}
A.~Koenecke.
\newblock {Racial Disparities in Automated Speech Recognition}.
\newblock In \emph{Proc. 2021 ISCA Symposium on Security and Privacy in Speech Communication}, 2021.

\bibitem[Koenecke et~al.(2020)Koenecke, Nam, Lake, Nudell, Quartey, Mengesha, Toups, Rickford, Jurafsky, and Goel]{koenecke2020racial}
A.~Koenecke, A.~Nam, E.~Lake, J.~Nudell, M.~Quartey, Z.~Mengesha, C.~Toups, J.~R. Rickford, D.~Jurafsky, and S.~Goel.
\newblock Racial disparities in automated speech recognition.
\newblock \emph{Proceedings of the National Academy of Sciences}, 117\penalty0 (14):\penalty0 7684--7689, 2020.

\bibitem[Lakshminarayanan et~al.(2017)Lakshminarayanan, Pritzel, and Blundell]{deepensemble}
B.~Lakshminarayanan, A.~Pritzel, and C.~Blundell.
\newblock Simple and scalable predictive uncertainty estimation using deep ensembles.
\newblock \emph{Advances in neural information processing systems}, 30, 2017.

\bibitem[Liu and Li(2023)]{liu2023understanding}
S.~Liu and X.~Li.
\newblock Understanding uncertainty sampling.
\newblock \emph{arXiv preprint arXiv:2307.02719}, 2023.

\bibitem[Mehrabi et~al.(2021)Mehrabi, Morstatter, Saxena, Lerman, and Galstyan]{uncertainty_fairness_3}
N.~Mehrabi, F.~Morstatter, N.~Saxena, K.~Lerman, and A.~Galstyan.
\newblock A survey on bias and fairness in machine learning.
\newblock \emph{ACM Comput. Surv.}, 54\penalty0 (6), jul 2021.
\newblock ISSN 0360-0300.
\newblock \doi{10.1145/3457607}.
\newblock URL \url{https://doi.org/10.1145/3457607}.

\bibitem[Mengesha et~al.(2021)Mengesha, Heldreth, Lahav, Sublewski, and Tuennerman]{mengesha2021don}
Z.~Mengesha, C.~Heldreth, M.~Lahav, J.~Sublewski, and E.~Tuennerman.
\newblock “i don’t think these devices are very culturally sensitive.”—impact of automated speech recognition errors on african americans.
\newblock \emph{Frontiers in Artificial Intelligence}, page 169, 2021.

\bibitem[Mitchell et~al.(2019)Mitchell, Wu, Zaldivar, Barnes, Vasserman, Hutchinson, Spitzer, Raji, and Gebru]{uncertainty_fairness_2}
M.~Mitchell, S.~Wu, A.~Zaldivar, P.~Barnes, L.~Vasserman, B.~Hutchinson, E.~Spitzer, I.~D. Raji, and T.~Gebru.
\newblock Model cards for model reporting.
\newblock In \emph{Proceedings of the Conference on Fairness, Accountability, and Transparency}, FAT* '19, page 220–229, New York, NY, USA, 2019. Association for Computing Machinery.
\newblock ISBN 9781450361255.
\newblock \doi{10.1145/3287560.3287596}.
\newblock URL \url{https://doi.org/10.1145/3287560.3287596}.

\bibitem[Nallasamy et~al.(2012)Nallasamy, Metze, and Schultz]{accent_al}
U.~Nallasamy, F.~Metze, and T.~Schultz.
\newblock Active learning for accent adaptation in automatic speech recognition.
\newblock In \emph{2012 IEEE Spoken Language Technology Workshop (SLT)}, pages 360--365, 2012.
\newblock \doi{10.1109/SLT.2012.6424250}.

\bibitem[Olatunji et~al.(2023{\natexlab{a}})Olatunji, Afonja, Dossou, Tonja, Emezue, Rufai, and Singh]{olatunji23_interspeech}
T.~Olatunji, T.~Afonja, B.~F.~P. Dossou, A.~L. Tonja, C.~C. Emezue, A.~M. Rufai, and S.~Singh.
\newblock {AfriNames: Most ASR Models "Butcher" African Names}.
\newblock In \emph{Proc. INTERSPEECH 2023}, pages 5077--5081, 2023{\natexlab{a}}.
\newblock \doi{10.21437/Interspeech.2023-2122}.

\bibitem[Olatunji et~al.(2023{\natexlab{b}})Olatunji, Afonja, Yadavalli, Emezue, Singh, Dossou, Osuchukwu, Osei, Tonja, Etori, and Mbataku]{Olatunji2023AfriSpeech200PA}
T.~Olatunji, T.~Afonja, A.~Yadavalli, C.~C. Emezue, S.~Singh, B.~F.~P. Dossou, J.~Osuchukwu, S.~Osei, A.~L. Tonja, N.~A. Etori, and C.~Mbataku.
\newblock Afrispeech-200: Pan-african accented speech dataset for clinical and general domain asr.
\newblock 2023{\natexlab{b}}.
\newblock URL \url{https://api.semanticscholar.org/CorpusID:263334123}.

\bibitem[Radford et~al.(2022)Radford, Kim, Xu, Brockman, McLeavey, and Sutskever]{whisper}
A.~Radford, J.~W. Kim, T.~Xu, G.~Brockman, C.~McLeavey, and I.~Sutskever.
\newblock Robust speech recognition via large-scale weak supervision, 2022.

\bibitem[Radford et~al.(2023)Radford, Kim, Xu, Brockman, McLeavey, and Sutskever]{radford2023robust}
A.~Radford, J.~W. Kim, T.~Xu, G.~Brockman, C.~McLeavey, and I.~Sutskever.
\newblock Robust speech recognition via large-scale weak supervision.
\newblock In \emph{International Conference on Machine Learning}, pages 28492--28518. PMLR, 2023.

\bibitem[Riccardi and Hakkani-Tur(2005)]{al_speech_2}
G.~Riccardi and D.~Hakkani-Tur.
\newblock Active learning: theory and applications to automatic speech recognition.
\newblock \emph{IEEE Transactions on Speech and Audio Processing}, 13\penalty0 (4):\penalty0 504--511, 2005.
\newblock \doi{10.1109/TSA.2005.848882}.

\bibitem[Selbst et~al.(2019)Selbst, Boyd, Friedler, Venkatasubramanian, and Vertesi]{uncertainty_fairness_1}
A.~D. Selbst, D.~Boyd, S.~A. Friedler, S.~Venkatasubramanian, and J.~Vertesi.
\newblock Fairness and abstraction in sociotechnical systems.
\newblock In \emph{Proceedings of the Conference on Fairness, Accountability, and Transparency}, FAT* '19, page 59–68, New York, NY, USA, 2019. Association for Computing Machinery.
\newblock ISBN 9781450361255.
\newblock \doi{10.1145/3287560.3287598}.
\newblock URL \url{https://doi.org/10.1145/3287560.3287598}.

\bibitem[Sener and Savarese(2017)]{sener2017active}
O.~Sener and S.~Savarese.
\newblock Active learning for convolutional neural networks: A core-set approach.
\newblock \emph{arXiv preprint arXiv:1708.00489}, 2017.

\bibitem[Settles(2009)]{settles2009active}
B.~Settles.
\newblock Active learning literature survey.
\newblock 2009.
\newblock URL \url{https://api.semanticscholar.org/CorpusID:324600}.

\bibitem[Tsvetkov(2017)]{tsvetkov2017opportunities}
Y.~Tsvetkov.
\newblock Opportunities and challenges in working with low-resource languages.
\newblock In \emph{Carnegie Mellon Univ., Language Technologies Institute}, 2017.

\bibitem[Yemmene and Besacier(2019)]{yemmene2019motivations}
P.~Yemmene and L.~Besacier.
\newblock Motivations, challenges, and perspectives for the development of an automatic speech recognition system for the under-resourced ngiemboon language.
\newblock In \emph{Proceedings of The First International Workshop on NLP Solutions for Under Resourced Languages (NSURL 2019) co-located with ICNLSP 2019-Short Papers}, pages 59--67, 2019.

\bibitem[Zapata and Kirkedal(2015)]{zapata2015assessing}
J.~Zapata and A.~S. Kirkedal.
\newblock Assessing the performance of automatic speech recognition systems when used by native and non-native speakers of three major languages in dictation workflows.
\newblock In \emph{Proceedings of the 20th Nordic Conference of Computational Linguistics (NODALIDA 2015)}, pages 201--210, 2015.

\end{thebibliography}
\appendix

\section{Appendices}

\subsection{Hyper-parameters}
\label{sec:para}
Table \ref{tab:paramter} shows the hyper-parameter settings used in this study. The top-k value in the table is changed according to the domain used in each of the experiments. For example, when conducting experiments in the general domain, we set the value of top-k to 2k.
\begin{table*}[ht!] 
\centering
\begin{tabular}{ll}
\hline
\textbf{Hyper-parameters} & \textbf{Values} \\
\hline
attention dropout & 0.1 \\
 hidden  dropout & 0.1 \\
layer drop & 0.1 \\
train batch size &  16 \\
val batch size &  8\\
number of epochs &  5 \\
learning rate & 3e-4 \\
maximum audio length &  260000 \\
maximum label length&  260 \\
minimum transcript length &  10 \\
top\_k & 2000, 3500, 6500 \\
domains & general, clinical, all \\
active learning rounds &  3 \\
sampling mode &  EU-Most, random \\
MC-Dropout round & 10\\\hline
\end{tabular}
\caption{Hyper-parameters summary}
\label{tab:paramter}
\end{table*}
\subsection{Country Statistics}\label{country}
Table \ref{tab:Country Stats} shows the countries' statistics across the AfriSpeech-200 dataset. 
\begin{table*}[ht!]
\small
\centering
\begin{tabular}{lllll}
\hline
\textbf{Country} & \textbf{Clips} & \textbf{Speakers} & \textbf{Duration (seconds)} & \textbf{Duration (hrs)} \\
\hline
Nigeria & 45875 &  1979 & 512646.88 & 142.40 \\
Kenya & 8304 &  137 &  75195.43 &  20.89 \\
South Africa &  7870 &  223 &  81688.11 &  22.69 \\
Ghana &  2018 &  37 &  18581.13 &  5.16 \\
Botswana &  1391 &  38 &  14249.01 &  3.96 \\
Uganda &  1092 &  26 &  10420.42 &  2.89 \\
Rwanda &  469 &  9 &  5300.99 &  1.47\\
 United States of America  &  219 &  5 &  1900.98 &  0.53 \\
Turkey &  66 &  1 &  664.01 &  0.18 \\
Zimbabwe &  63 &  3 &  635.11 &  0.18 \\
Malawi &  60 &  1 &  554.61 & 0.15\\
Tanzania  &  51& 2 &  645.51 &  0.18 \\
Lesotho & 7 & 1 &  78.40 &  0.02\\ \hline
\end{tabular}
\caption{Countries Statistics across the dataset}
\label{tab:Country Stats}
\end{table*}

\subsection{Dataset Accents Stats}
Tables \ref{tab:dataset_accent_stats1} and \ref{tab:dataset_accent_stats2} provide a list of AfriSpeech accents along with the number of unique speakers, countries where speakers for each accent are located, duration in seconds for each accent, and their presence in the train, dev, and test splits.

\begin{table*}[ht!]
\small
\centering
\begin{tabular}{llllll}
\hline 
\textbf{Accent} & \textbf{Clips} & \textbf{Speakers}  & \textbf{Duration(s)} & \textbf{
Countries} & \textbf{Splits}\\
\hline
 yoruba & 15407 & 683 & 161587.55 & US,NG & train,test,dev \\ 
 igbo & 8677 & 374 & 93035.79 & US,NG,ZA & train,test,dev \\ 
 swahili & 6320 & 119 & 55932.82 & KE,TZ,ZA,UG & train,test,dev \\ 
 hausa & 5765 & 248 & 70878.67 & NG & train,test,dev \\ 
 ijaw & 2499 & 105 & 33178.9 & NG & train,test,dev \\ 
 afrikaans & 2048 & 33 & 20586.49 & ZA & train,test,dev \\ 
 idoma & 1877 & 72 & 20463.6 & NG & train,test,dev \\ 
 zulu & 1794 & 52 & 18216.97 & ZA,TR,LS & dev,train,test \\ 
 setswana & 1588 & 39 & 16553.22 & BW,ZA & dev,test,train \\ 
 twi & 1566 & 22 & 14340.12 & GH & test,train,dev \\ 
 isizulu & 1048 & 48 & 10376.09 & ZA & test,train,dev \\ 
 igala & 919 & 31 & 9854.72 & NG & train,test \\ 
 izon & 838 & 47 & 9602.53 & NG & train,dev,test \\ 
 kiswahili & 827 & 6 & 8988.26 & KE & train,test \\ 
 ebira & 757 & 42 & 7752.94 & NG & train,test,dev \\ 
 luganda & 722 & 22 & 6768.19 & UG,BW,KE & test,dev,train \\ 
 urhobo & 646 & 32 & 6685.12 & NG & train,dev,test \\ 
 nembe & 578 & 16 & 6644.72 & NG & train,test,dev \\ 
 ibibio & 570 & 39 & 6489.29 & NG & train,test,dev \\ 
 pidgin & 514 & 20 & 5871.57 & NG & test,train,dev \\ 
 luhya & 508 & 4 & 4497.02 & KE & train,test \\ 
 kinyarwanda & 469 & 9 & 5300.99 & RW & train,test,dev \\ 
 xhosa & 392 & 12 & 4604.84 & ZA & train,dev,test \\ 
 tswana & 387 & 18 & 4148.58 & ZA,BW & train,test,dev \\ 
 esan & 380 & 13 & 4162.63 & NG & train,test,dev \\ 
 alago & 363 & 8 & 3902.09 & NG & train,test \\ 
 tshivenda & 353 & 5 & 3264.77 & ZA & test,train \\ 
 fulani & 312 & 18 & 5084.32 & NG & test,train \\ 
 isoko & 298 & 16 & 4236.88 & NG & train,test,dev \\ 
 akan (fante) & 295 & 9 & 2848.54 & GH & train,dev,test \\ 
 ikwere & 293 & 14 & 3480.43 & NG & test,train,dev \\ 
 sepedi & 275 & 10 & 2751.68 & ZA & dev,test,train \\ 
 efik & 269 & 11 & 2559.32 & NG & test,train,dev \\ 
 edo & 237 & 12 & 1842.32 & NG & train,test,dev \\ 
 luo & 234 & 4 & 2052.25 & UG,KE & test,train,dev \\ 
 kikuyu & 229 & 4 & 1949.62 & KE & train,test,dev \\ 
 bekwarra & 218 & 3 & 2000.46 & NG & train,test \\ 
 isixhosa & 210 & 9 & 2100.28 & ZA & train,dev,test \\ 
 hausa/fulani & 202 & 3 & 2213.53 & NG & test,train \\ 
 epie & 202 & 6 & 2320.21 & NG & train,test \\ 
 isindebele & 198 & 2 & 1759.49 & ZA & train,test \\ 
 venda and xitsonga & 188 & 2 & 2603.75 & ZA & train,test \\ 
 sotho & 182 & 4 & 2082.21 & ZA & dev,test,train \\ 
 akan & 157 & 6 & 1392.47 & GH & test,train \\ 
 nupe & 156 & 9 & 1608.24 & NG & dev,train,test \\ 
 anaang & 153 & 8 & 1532.56 & NG & test,dev \\ 
 english & 151 & 11 & 2445.98 & NG & dev,test \\ 
 afemai & 142 & 2 & 1877.04 & NG & train,test \\ 
 shona & 138 & 8 & 1419.98 & ZA,ZW & test,train,dev \\ 
 eggon & 137 & 5 & 1833.77 & NG & test \\ 
 luganda and kiswahili & 134 & 1 & 1356.93 & UG & train \\ 
 ukwuani & 133 & 7 & 1269.02 & NG & test \\ 
 sesotho & 132 & 10 & 1397.16 & ZA & train,dev,test \\ 
 benin & 124 & 4 & 1457.48 & NG & train,test \\ 
 kagoma & 123 & 1 & 1781.04 & NG & train \\ 
 nasarawa eggon & 120 & 1 & 1039.99 & NG & train \\ 
 tiv & 120 & 14 & 1084.52 & NG & train,test,dev \\ 
 south african english & 119 & 2 & 1643.82 & ZA & train,test \\ 
 borana & 112 & 1 & 1090.71 & KE & train \\ 
\hline
\end{tabular}
\caption{Dataset Accent Stats, Part I}
\label{tab:dataset_accent_stats1}
\end{table*}

\begin{table*}[ht!]
\small
\centering
\begin{tabular}{llllll}
\hline 
\textbf{Accent} & \textbf{Clips} & \textbf{Speakers}  & \textbf{Duration(s)} & \textbf{
Countries} & \textbf{Splits}\\
\hline

 swahili ,luganda ,arabic & 109 & 1 & 929.46 & UG & train \\ 
 ogoni & 109 & 4 & 1629.7 & NG & train,test \\ 
 mada & 109 & 2 & 1786.26 & NG & test \\ 
 bette & 106 & 4 & 930.16 & NG & train,test \\ 
 berom & 105 & 4 & 1272.99 & NG & dev,test \\ 
 bini & 104 & 4 & 1499.75 & NG & test \\ 
 ngas & 102 & 3 & 1234.16 & NG & train,test \\ 
 etsako & 101 & 4 & 1074.53 & NG & train,test \\ 
 okrika & 100 & 3 & 1887.47 & NG & train,test \\ 
 venda & 99 & 2 & 938.14 & ZA & train,test \\ 
 siswati & 96 & 5 & 1367.45 & ZA & dev,train,test \\ 
 damara & 92 & 1 & 674.43 & NG & train \\ 
 yoruba, hausa & 89 & 5 & 928.98 & NG & test \\ 
 southern sotho & 89 & 1 & 889.73 & ZA & train \\ 
 kanuri & 86 & 7 & 1936.78 & NG & test,dev \\ 
 itsekiri & 82 & 3 & 778.47 & NG & test,dev \\ 
 ekpeye & 80 & 2 & 922.88 & NG & test \\ 
 mwaghavul & 78 & 2 & 738.02 & NG & test \\ 
 bajju & 72 & 2 & 758.16 & NG & test \\ 
 luo, swahili & 71 & 1 & 616.57 & KE & train \\ 
 dholuo & 70 & 1 & 669.07 & KE & train \\ 
 ekene & 68 & 1 & 839.31 & NG & test \\ 
 jaba & 65 & 2 & 540.66 & NG & test \\ 
 ika & 65 & 4 & 576.56 & NG & test,dev \\ 
 angas & 65 & 1 & 589.99 & NG & test \\ 
 ateso & 63 & 1 & 624.28 & UG & train \\ 
 brass & 62 & 2 & 900.04 & NG & test \\ 
 ikulu & 61 & 1 & 313.2 & NG & test \\ 
 eleme & 60 & 2 & 1207.92 & NG & test \\ 
 chichewa & 60 & 1 & 554.61 & MW & train \\ 
 oklo & 58 & 1 & 871.37 & NG & test \\ 
 meru & 58 & 2 & 865.07 & KE & train,test \\ 
 agatu & 55 & 1 & 369.11 & NG & test \\ 
 okirika & 54 & 1 & 792.65 & NG & test \\ 
 igarra & 54 & 1 & 562.12 & NG & test \\ 
 ijaw(nembe) & 54 & 2 & 537.56 & NG & test \\ 
 khana & 51 & 2 & 497.42 & NG & test \\ 
 ogbia & 51 & 4 & 461.15 & NG & test,dev \\ 
 gbagyi & 51 & 4 & 693.43 & NG & test \\ 
 portuguese & 50 & 1 & 525.02 & ZA & train \\ 
 delta & 49 & 2 & 425.76 & NG & test \\ 
 bassa & 49 & 1 & 646.13 & NG & test \\ 
 etche & 49 & 1 & 637.48 & NG & test \\ 
 kubi & 46 & 1 & 495.21 & NG & test \\ 
 jukun & 44 & 2 & 362.12 & NG & test \\ 
 igbo and yoruba & 43 & 2 & 466.98 & NG & test \\ 
 urobo & 43 & 3 & 573.14 & NG & test \\ 
 kalabari & 42 & 5 & 305.49 & NG & test \\ 
 ibani & 42 & 1 & 322.34 & NG & test \\ 
 obolo & 37 & 1 & 204.79 & NG & test \\ 
 idah & 34 & 1 & 533.5 & NG & test \\ 
 bassa-nge/nupe & 31 & 3 & 267.42 & NG & test,dev \\ 
 yala mbembe & 29 & 1 & 237.27 & NG & test \\ 
 eket & 28 & 1 & 238.85 & NG & test \\ 
 afo & 26 & 1 & 171.15 & NG & test \\ 
 ebiobo & 25 & 1 & 226.27 & NG & test \\ 
 nyandang & 25 & 1 & 230.41 & NG & test \\ 
 ishan & 23 & 1 & 194.12 & NG & test \\ 
 bagi & 20 & 1 & 284.54 & NG & test \\ 
 estako & 20 & 1 & 480.78 & NG & test \\ 
 gerawa & 13 & 1 & 342.15 & NG & test \\ 
\hline
\end{tabular}
\caption{Dataset Accent Stats, Part II}
\label{tab:dataset_accent_stats2}
\end{table*}
 \subsection{Most common accent distribution}
\begin{figure*}[!ht]
      \centering
\includegraphics[width=\linewidth]{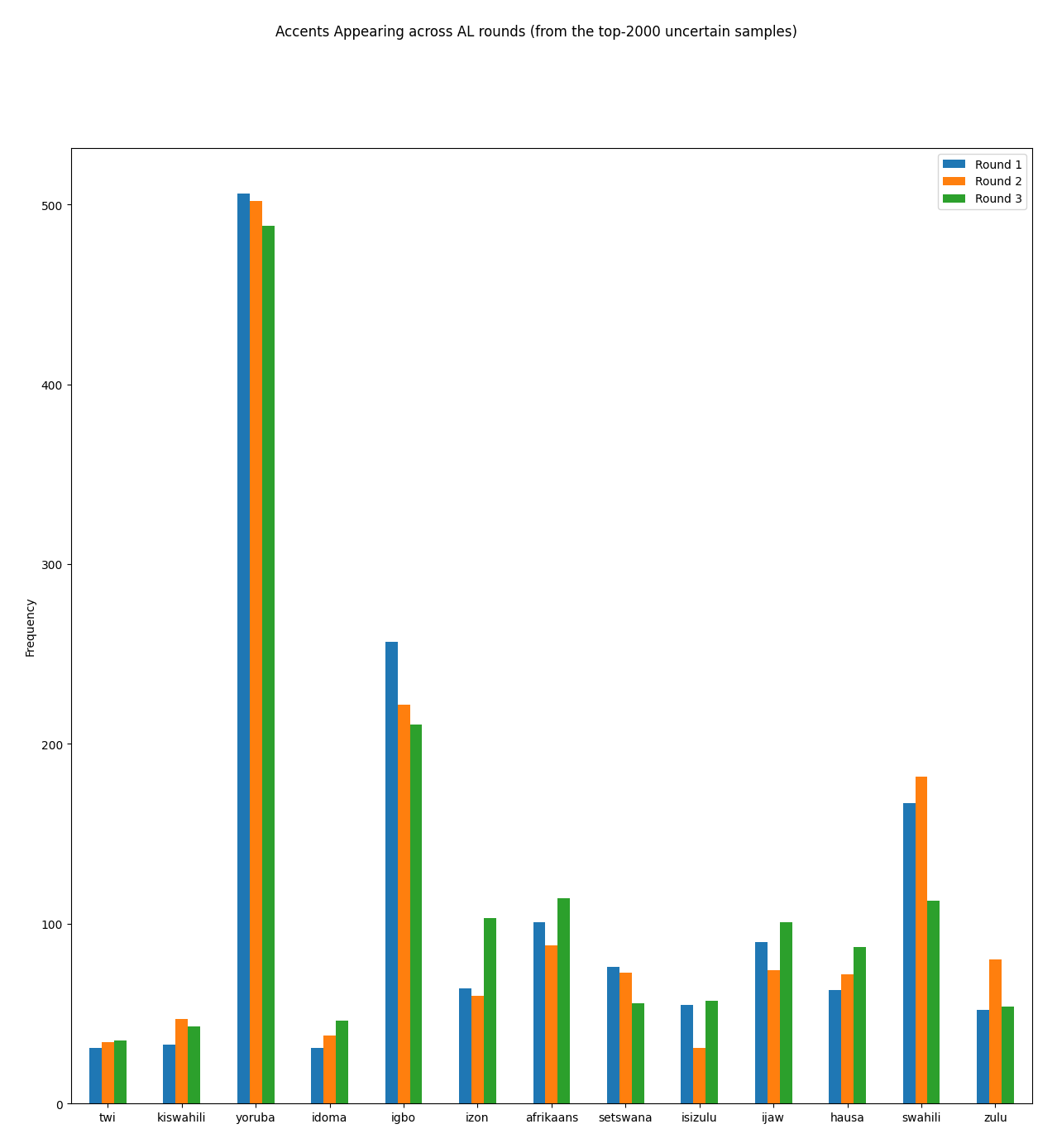}
     \caption{Most common accents distribution across the general domain with EU-Most sampling strategy.}
     \label{fig:mostd}
 \end{figure*}

\begin{figure*}[ht!]
     \centering
\includegraphics[width=\linewidth]{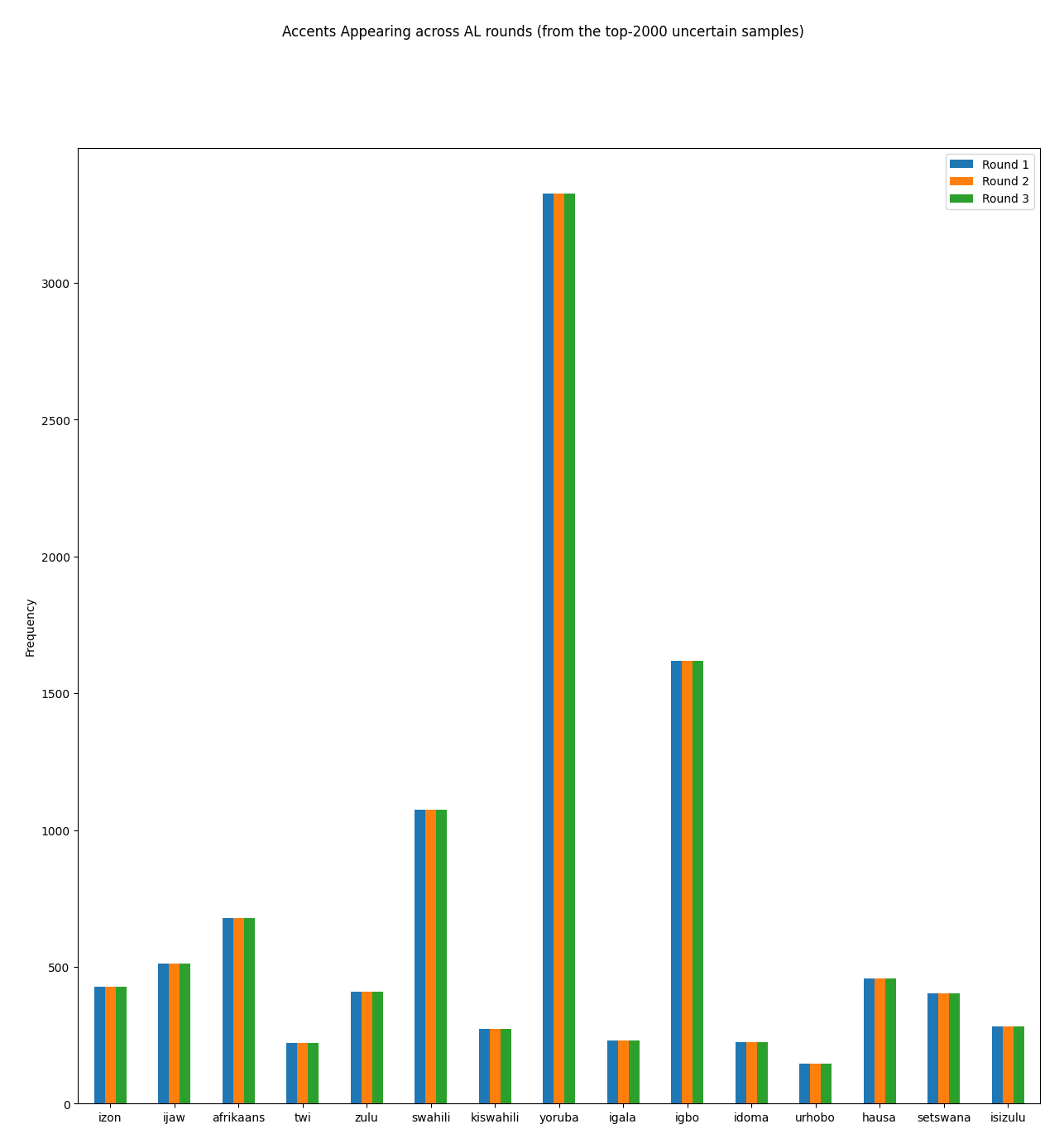}
     \caption{Most common accents distribution across the general domain with random selection strategy.}
     \label{fig:randd}
 \end{figure*}

 Figures \ref{fig:mostd} and \ref{fig:randd} show the most common accent distribution across the general domain with random and EU-Most selection strategies.
 \subsection{Ascending and Descending Accents}
\begin{figure*}[ht!]
     \centering
     \includegraphics[width=\linewidth]{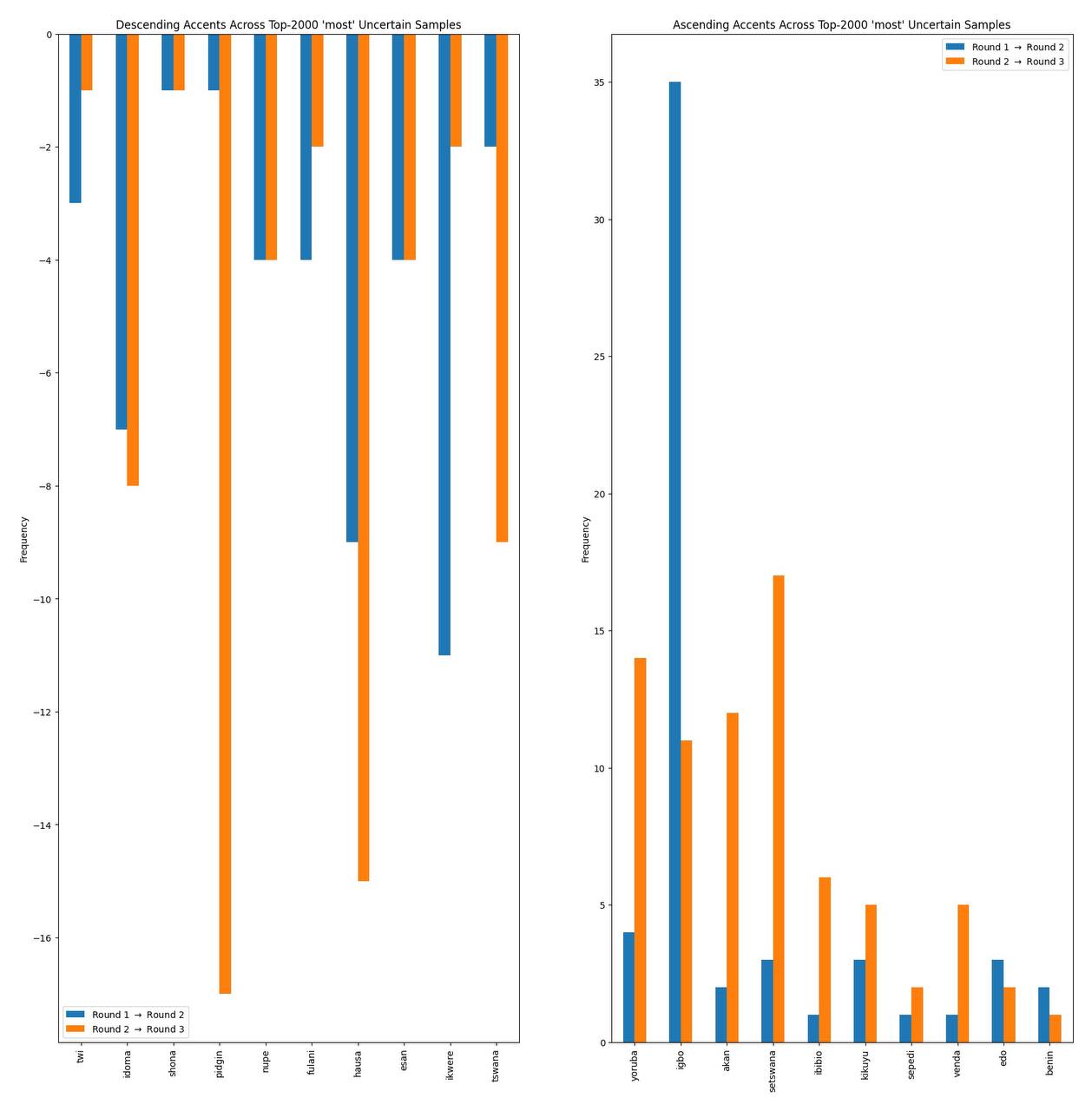}
     \caption{Ascending and descending accents across Top-2K \textit{most} uncertain samples.}
     \label{fig:dec}
 \end{figure*}

 Figure \ref{fig:dec} shows ascending and descending accents across the Top 2k \textit{most} uncertain samples.
\end{document}